\documentclass{article}

\usepackage{microtype}
\usepackage{graphicx}
\usepackage{subfigure}
\usepackage{booktabs, makecell} 
\usepackage[hyphens]{url}
\usepackage{textcomp}
\usepackage{multirow}
\usepackage{tabulary}
\usepackage{amsmath, amssymb, amsfonts}
\usepackage{threeparttable}
\usepackage{natbib}

\usepackage{hyperref}

\usepackage[accepted]{icml2020}

\icmltitlerunning{Learning Multiplicative Interactions with Bayesian Neural Networks for Visual-Inertial Odometry}

\begin{document}

\twocolumn[
\icmltitle{Learning Multiplicative Interactions with Bayesian \\ Neural Networks for Visual-Inertial Odometry}



\begin{icmlauthorlist}
\icmlauthor{Kashmira Shinde}{goo}
\icmlauthor{Jongseok Lee}{goo}
\icmlauthor{Matthias Humt}{goo}
\icmlauthor{Aydin Sezgin}{to}
\icmlauthor{Rudolph Triebel}{goo}
\end{icmlauthorlist}

\icmlaffiliation{to}{Institute of Digital Communication Systems, Ruhr-Universität Bochum (RUB), Germany}
\icmlaffiliation{goo}{German Aerospace Center (DLR), Institute of Robotics and Mechatronics,  Wessling, Germany}

\icmlcorrespondingauthor{Kashmira Shinde}{kashmira.shinde23@gmail.com}

\icmlkeywords{Deep Learning, Sensor Fusion, Multi-head Attention, Uncertainty Estimation, Laplace Approximation}

\vskip 0.3in
]



\printAffiliationsAndNotice{} 

\begin{abstract}
This paper presents an end-to-end multi-modal learning approach for monocular Visual-Inertial Odometry (VIO), which is specifically designed to exploit sensor complementarity in the light of sensor degradation scenarios. The proposed network makes use of a multi-head self-attention mechanism that learns multiplicative interactions between multiple streams of information. Another design feature of our approach is the incorporation of the model uncertainty using scalable Laplace Approximation. We evaluate the performance of the proposed approach by comparing it against the end-to-end state-of-the-art methods on the KITTI dataset and show that it achieves superior performance. Importantly, our work thereby provides an empirical evidence that learning multiplicative interactions can result in a powerful inductive bias for increased robustness to sensor failures.
\end{abstract}

\section{Introduction}
\label{intro}
For safety critical applications such as autonomous driving, it is crucial for learning approaches to effectively rely on multiple sources of information. In case of a localization problem as an example, learning from complementary sensor types such as cameras and IMUs can bring improved robustness and redundancy when there exists sensor corruptions and failures. Existing methods for learning based VIO rely on different strategies to fuse visual and inertial data, which include so-called \textit{FC-fusion networks} \citep{vinet, deepvio} and \textit{soft/hard feature selection networks} \citep{ssf, selfvio}. Broadly speaking, these strategies typically learn additive interactions for integrating multiple streams of information, i.e. , given the features $\textbf{u}$ and $\textbf{v}$ from different modalities, the additive interactions are modeled as $f(\textbf{u}, \textbf{v}) = \textbf{W}\texttt{concat}(\textbf{u},\textbf{v})$ where $\texttt{concat}(\textbf{u},\textbf{v})$ is the concatenation of the two features, and $\textbf{W}$ are the learned fusion network parameters.

Instead, we present an end-to-end trainable deep neural network for monocular VIO with the focus on how to better exploit sensor complementarity. Specifically, the designed network architecture (i) learns multiplicative interactions \citep{multiplicative} for the fusion network, which is based on multi-head self-attention layers (section \ref{fusion}), and (ii) treat the network weights as probability distribution using scalable \textit{Laplace Approximation} (LA) \citep{mackay1992practical}, resulting in \textit{Bayesian Neural Networks} (Section \ref{ua}). In our experiments, we compete with the current methods that learn additive interactions against the proposed network (section \ref{res1}) and further evaluate the obtained uncertainty (section \ref{res2}). As a result, we demonstrate improved performance in terms of accuracy and robustness. We believe that these results imply that a more elaborated fusion might bring performance improvements for the end-to-end VIO.

\begin{figure*}[ht]
  \centering
  \includegraphics[width=15cm, height=8cm]{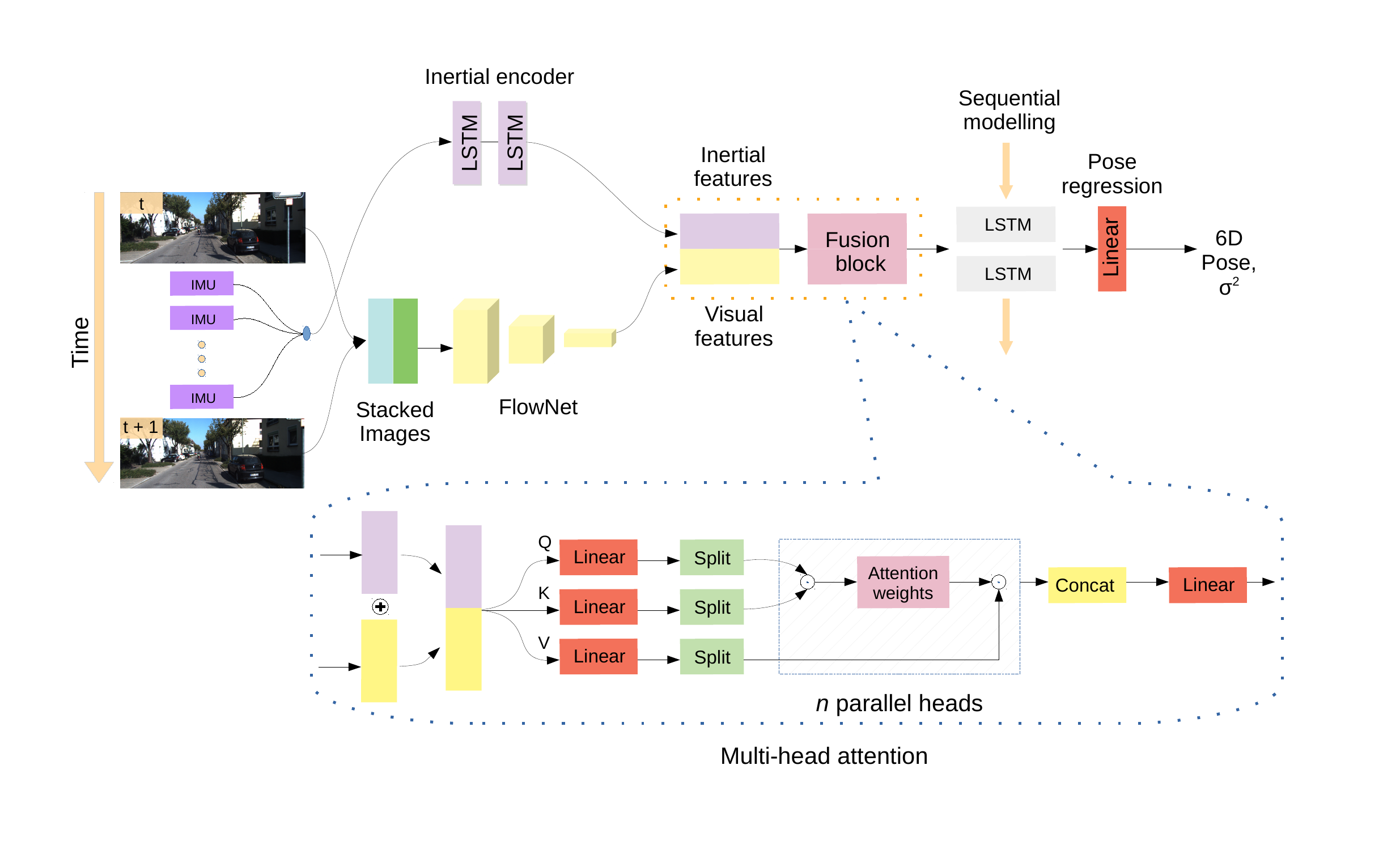}
  \vspace{-5ex}
  \caption{Architecture of the proposed approach. $\sigma^2$ denotes the variance, i.e., uncertainty of the poses.}
  \label{fig:arch}
\end{figure*}

\section{Method}
We now describe the proposed architecture in figure \ref{fig:arch}.

\subsection{Visual-Inertial Feature Extractors}
Two sequential images, i.e., \textbf{u}\textit{\textsubscript{v1}}, \textbf{u}\textit{\textsubscript{v2}} are stacked together and fed to the optical flow feature extractor (FlowNet \citet{flownet} or even \citet{sun2018pwc}) \textit{g}\text{\textsubscript{vision}}. It learns geometric feature representation to be able to generalise well in unfamiliar environments as opposed to learning appearance representation. The visual features \textbf{b}\textit{\textsubscript{v}}  obtained from the final layer of FlowNet can be expressed as:
\begin{equation}
  \begin{aligned}
    \textbf{b}\textit{\textsubscript{v}} &=  g_{\text{vision}}(\textbf{u}\textit{\textsubscript{v1}}, \textbf{u}\textit{\textsubscript{v2}}). \\
  \end{aligned}
\end{equation}

Raw inertial measurements \textbf{u}\textit{\textsubscript{i}} falling between two consequent image frames, i.e., $x, y, z$ components of linear and angular acceleration taken together forming a 6 dimensional vector are passed to an encoder  \textit{g}\text{\textsubscript{inertial}}. It is composed of a bidirectional LSTM (Long Short-Term Memory) as done by \cite{ionet} and operates at a rate at which IMU receives data. The inertial features \textbf{b}\textit{\textsubscript{i}} can be expressed as:
\begin{equation}
  \begin{aligned}
    \textbf{b}\textit{\textsubscript{i}} &=  g_{\text{inertial}}(\textbf{u}\textit{\textsubscript{i}}). \\
  \end{aligned}
\end{equation}

\subsection{Feature Fusion Mechanism}
\label{fusion}
The fusion block combines both the inertial and visual feature streams generated from the respective feature extractors. Here, a widely used fusion method is to either simply concatenate the two feature streams into a single feature space \cite{vinet} or re-weight them \cite{ssf}, which are so-called additive interactions. In other words, given two different input modalities $\textbf{u}$ and $\textbf{v}$, the used fusion blocks rely on rather simple functions: ($\texttt{concat}(\textbf{u},\textbf{v})$) or a linear layer $f(\textbf{u},\textbf{v}) =  \textbf{W}\texttt{concat}(\textbf{u},\textbf{v}) + \textbf{a}$.

In contrast to this, we consider so-called \textit{multiplicative interactions} (MI). For deep neural networks, MI can be expressed by the following function \citep{multiplicative}:
\begin{equation}
  \label{eq:mul}
  \begin{aligned}
    f(\textbf{u},\textbf{v}) &=  \textbf{v}^{\top}\mathcal{W}\textbf{u} + \textbf{v}^{\top}\textbf{W}_{v} + \textbf{W}_{u}\textbf{u} + \textbf{a}, \\
  \end{aligned}
\end{equation}
wherein the weight matrices $\textbf{W}_{u}$ and $\textbf{W}_{v}$, vector $\textbf{a}$ and 3D weight tensor $\mathcal{W}$ are the learned parameters. A major difference to additive interaction is the product term $\textbf{v}^{\top}\mathcal{W}\textbf{u}$. Here, one might argue that neural networks are universal approximators, and thus, adding more costly module may not affect the approximation power of neural networks. Yet, the hypothesis space or the sets of learnable functions, the compactness of the estimator as well as learnability can increase as proved by \citet{multiplicative}. When learned correctly, MI can help in preferential selection of hypotheses, and thus may result in an improved inductive bias.

Motivated by this recent result, we proposed to use a more elaborated fusion strategy that can learn MI. To this end, we propose a fusion block based on multi-head self-attention (denoted by $f_\text{attn}$) \cite{attneed} which relates different parts of the same sequence to learn a suitable combined feature representation. An advantage over other MI modules is that training can be often easier (e.g. alternatives such as gating networks are often notoriously difficult to train \citep{Shazeer17}). In this module, the concatenated visual and inertial features of dimension $d_{k}$ are passed to the attention function as query $\mathbf{q}$ and a key-value pair ($\mathbf{k}$,$\mathbf{v}$). In order to learn distinct representations of the input, $n$ such attention \textit{heads} perform attention function $\mathcal{A}$ in parallel for a model of dimension $N$. The output from each head $\mathbf{h_i}$ is then concatenated to produce final attention outputs for a given concatenated feature representation:
\begin{equation}
    f_\text{attn}(\mathbf{q}, \mathbf{k}, \mathbf{v}) = \mathbf{h}\mathbf{W^{h}},
\end{equation}

where each head $\mathbf{h_i}$ for $i=1,...,n$ is given by
\begin{align*}
    \mathbf{h_i} &= \mathcal{A}(\mathbf{q}\mathbf{W_{i}^q}, \mathbf{k}\mathbf{W_{i}^k}, \mathbf{v}\mathbf{W_{i}^v}) \\
    \mathcal{A}(\mathbf{q}, \mathbf{k}, \mathbf{v}) &= \texttt{softmax}\left(\frac{\mathbf{q} \mathbf{k}^{\top}}{\sqrt{d_{k}}}\right)\mathbf{v}, \\
\label{eq:def:mha}
\end{align*}
where $\mathbf{h} = \texttt{concat}(\mathbf{h_1}, ..., \mathbf{h_n})$ and learned attention weights $\mathbf{W_{i}^q} \in \mathbb{R}^{N \times d_{k}}$ , $\mathbf{W_{i}^k} \in  \mathbb{R}^{N \times d_{k}}$ , $\mathbf{W_{i}^v} \in \mathbb{R}^{N \times d_{k}}$ and $\mathbf{W^{h}} \in  \mathbb{R}^{nd_{k} \times N}$. The attention outputs \textbf{y} obtained from the visual \textbf{b}\textit{\textsubscript{v}} and inertial \textbf{b}\textit{\textsubscript{i}} feature vectors which are passed to pose regression module can be expressed as:
\begin{equation}
  \begin{aligned}
    \textbf{y} &=  f_\text{attn}(\textbf{b}\textit{\textsubscript{v}},\textbf{b}\textit{\textsubscript{i}}). \\
  \end{aligned}
\end{equation}
We note that, while the used module is more costly than other simpler variants, we did not find training more difficult. Furthermore, we use the attention mechanism only in the fusion block, and so, the added number of parameters has smaller contribution to the total number of parameters.

\subsection{Sequential Modeling and Pose Regression}
Next, we describe modeling sequential dependence for ego-motion estimation. In order to learn complex model dynamics (motion model) and to derive connections between sequential features, the core LSTM is used. The resultant representation from the feature fusion block \textbf{y}\textit{\textsubscript{t}} at time $t$ is fed to the core LSTM along with its hidden states from the previous time step $\textbf{h}_{t-1}$. The fully connected layer does the pose regression and the output is a 6D camera pose $\textbf{x}_{t}$:
\begin{equation}
  \begin{aligned}
    \textbf{x}_{t} &=  f_\text{lstm}(\textbf{y}\textit{\textsubscript{t}},\textbf{h}_{t-1}). \\
  \end{aligned}
\end{equation}

\subsection{Loss Function}
The loss as in \citet{deepvo} constitutes of the summed up mean squared error (MSE) between predicted poses $\hat{\textbf{x}_{t}}$ = ($\hat{\textbf{z}}\textit{\textsubscript{t}}$,$\hat{\mathbf{\psi}}\textit{\textsubscript{t}}$) and ground truth $\textbf{x}_{t}$ = (\textbf{z}\textit{\textsubscript{t}},$\mathbf{\psi\textit{\textsubscript{t}}}$) \citep{kitti1} where \textbf{z}\textit{\textsubscript{t}} denotes position and $\mathbf{\psi\textit{\textsubscript{t}}}$ the orientation at time $t$,
  \begin{equation}
  \label{eq:mseloss}
    \ell = \frac{1}{M}\sum_{i=1}^{M}\sum_{t=1}^{k}\left\|\hat{\textbf{z}\textit{\textsubscript{t}}} - \textbf{z}\textit{\textsubscript{t}} \right\|^{2}_{2} + \beta\left\|\hat{\mathbf{\psi_{t}}} - \mathbf{\psi_{t}} \right\|^{2}_{2},
  \end{equation}
  
where $\beta$ is a scale factor to balance the position and orientation elements for $M$ mini-batch samples. This scaling factor is to balance the magnitude so that the both translation and rotation are given the similar level of importance. 

\subsection{Uncertainty estimation} \label{ua}
We estimate the predictive uncertainty of our method using a LA similar to \citet{ritter2018scalable, lee2020estimating}, but only make use of the diagonal approximation to the Fisher information matrix. Just like MC-dropout \cite{gal2016dropout} it is a practical method for uncertainty estimation in deep learning without the need to change or retrain the model, but without the reliance on dropout layers.

LA employs information from the inverse curvature (Hessian) of the loss for the covariance matrix of a multivariate Gaussian as an approximation to the model's weight posterior distribution. We collect the output of multiple stochastic forward passes obtained through sampling from this posterior and perform Bayesian model averaging using Monte Carlo integration to make predictions. Note that we do not consider aleatoric uncertainty as we focus on the influence of model uncertainty on deep sensor fusion. More details can be found in the supplementary material. 

\section{Experimental Results}
In this section, we evaluate our approach in comparison to the baselines - DeepVO \cite{deepvo}, VINet \cite{vinet} and VIO Soft \cite{ssf}. Implementation details can be found in supplementary material. 

\subsection{Analysis of Robustness to Sensor Degradation} \label{res1}

\begin{figure}[ht]
\centering     
\subfigure[Seq 10]{\includegraphics[width=40.5mm]{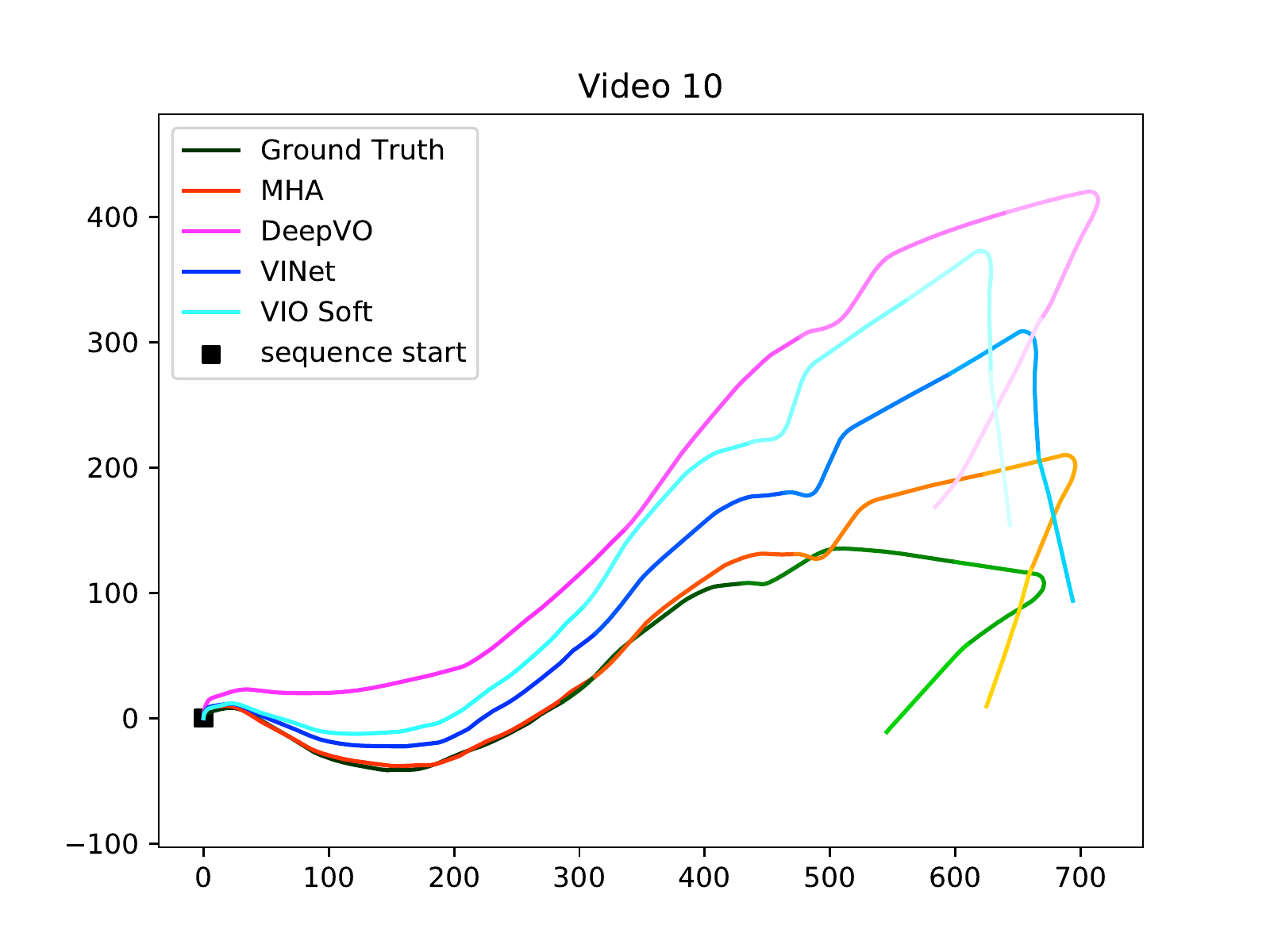}}
\subfigure[Seq 10 with vision deg.]{\includegraphics[width=40.5mm]{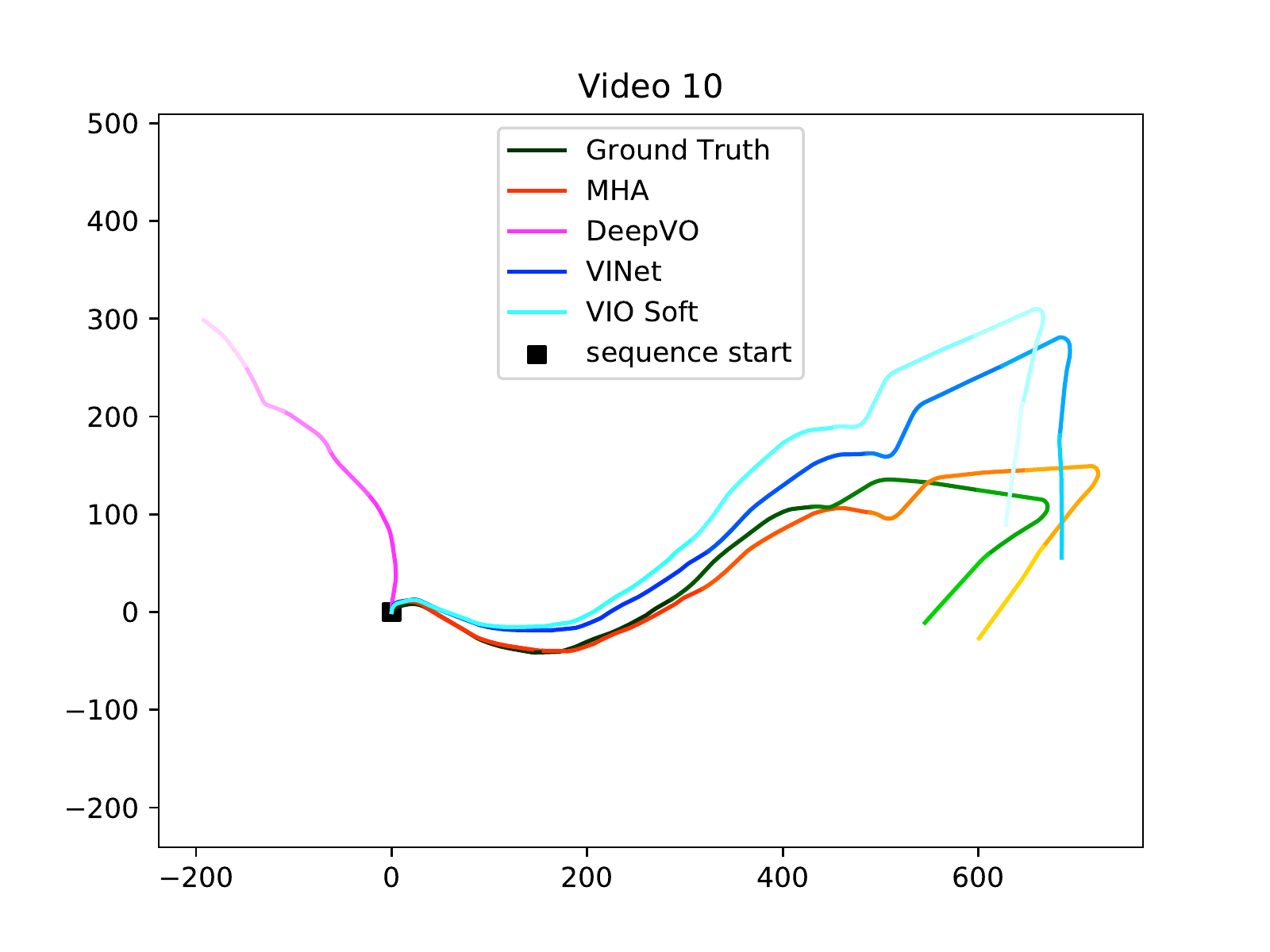}}
\caption[Caption for LOF]{Predicted trajectories of the KITTI dataset comparing MHA and baselines with ground truth (in green). (a) for non-degraded data. (b) for degraded vision data (occlusion, noise+blur, missing images) for sequence 10. Best viewed in colour.}\label{fig:traj}
\end{figure}

\begin{table*}[ht]
\begin{center}
\caption{Comparison of approaches for various sensor corruptions are reported along with nominal case. For complexity comparison, the total number of parameters in the networks and fusion blocks are shown. Per sequence results can be found in supplementary material.}
\vspace{1ex}
\begin{small}
\begin{sc}
\begin{tabulary}{\linewidth}{CCCCCCCCCCC}
\toprule
\multirow{2}{*}{Model} & \multirow{2}{*}{\#Total Param} & \multirow{2}{*}{ \#Fusion param} & \multicolumn{2}{c}{Nominal} & \multicolumn{2}{c}{Inertial} & \multicolumn{2}{c}{Vision} & \multicolumn{2}{c}{All} \\
\cmidrule(l){4-5}\cmidrule(l){6-7}\cmidrule(l){8-9}\cmidrule(l){10-11} & & & {$t_{rel}$} & {$r_{rel}$} & {$t_{rel}$} & {$r_{rel}$} & {$t_{rel}$} & {$r_{rel}$}  & {$t_{rel}$} & {$r_{rel}$} \\
\midrule
DeepVO & 149,518,326 & N/A & 18.07 & 5.88 & N/A & N/A & Fails & Fails &N/A&N/A\\
VINet & 294,828,726 & \textbf{31,020} & 13.98 & 4.66 & 16.31 & 5.28 & 15.46 & 4.97 &15.87&5.14\\
VIO Soft & \textbf{58,969,014} & 262,656 & 13.65 & 4.83 & 14.05 & 4.94 & 14.59 & 4.58 &15.48&4.63\\
MHA & 60,019,638 & 1,313,280 & \textbf{11.42} & \textbf{3.94} & \textbf{12.48} & \textbf{3.93} & \textbf{12.61} & \textbf{3.98} &\textbf{12.47}&\textbf{3.79}\\
\bottomrule
\label{table:nom}
\end{tabulary}
\end{sc}
\end{small}
\end{center}
\vspace{-3ex}
\begin{tablenotes}
  \centering
  \footnotesize
  \item $t_{rel}(\%)$: translational RMSE, $r_{rel}$(\textdegree/100m): rotational RMSE.
\end{tablenotes}
\end{table*}

We first compare the baselines with our approach (referred to as MHA) for nominal case where dataset has no degradation. In many real world scenarios, it is plausible for sensors to be imperfect thereby resulting in corrupted data. For a camera, these corruptions can be in the form of occlusions, noisy and blurry images as well as missing image frames. The IMU can have bias in gyroscope data as well as white noise in accelerometer data. To determine the behaviour of the approaches in such scenarios and evaluate the influence of the modalities, we prepare three datasets - a dataset with degraded inertial data, a vision degraded dataset (consisting of image occlusions, blur and missing data) and all sensor dataset (both vision and inertial data are degraded). 

Table \ref{table:nom} shows mean values of the sequences while its expanded version can be found in supplementary material. We use the metric provided by \citet{violearner} - average translation RMSE $t_{rel}(\%)$ and average rotational RMSE $r_{rel}$(\textdegree/100m) computed on sub-sequences of length 100m - 800m. In our experiments, we find that our method makes predictions with lower RMSE on average, when compared the considered baselines across the sequences. For inertial degradation, it can be seen that as the sensor fusion strategy becomes elaborate for each approach, the robustness to IMU degradation increases. For low to moderate intensity of degradation, the performance of all methods remains similar to nominal case, suggesting visual features are more dominant than inertial ones. For vision degradation case as the dataset is severely degraded, from Figure~\ref{fig:traj} we see that the performance of DeepVO degrades immensely and tracking is lost completely. Interestingly, when vision degrades, the translation error increases for all the approaches, but the rotational error remains comparable to the nominal case. This suggests that the visual features might contribute to determining translation and inertial contribute to rotation, and furthermore, inertial features might become more reliable in case of strong visual degradation.

To summarize, exploiting the combination of sensors can be beneficial for odometry in comparison to use of a single sensor (DeepVO). Among all three sensor fusion approaches, our approach being more expressive than VINet and VIO Soft, performs better than them on average in terms of accuracy (lower translational and orientation error) in our experiments. Our method appears to be more robust to sensor corruptions and tends to diverge less as compared to the other two approaches. We further report total number of parameters in a network, and also in the fusion block. As expected, our approach shows an increased cost for the fusion parameters, but requires not as much total network parameters to obtain an improved performance, suggesting the competitiveness of MI w.r.t performance vs cost trade-off.
\subsection{Results of the Bayesian Framework} \label{res2}
\begin{figure}[ht]
\centering     
\subfigure[Nominal]{\includegraphics[width=40.5mm]{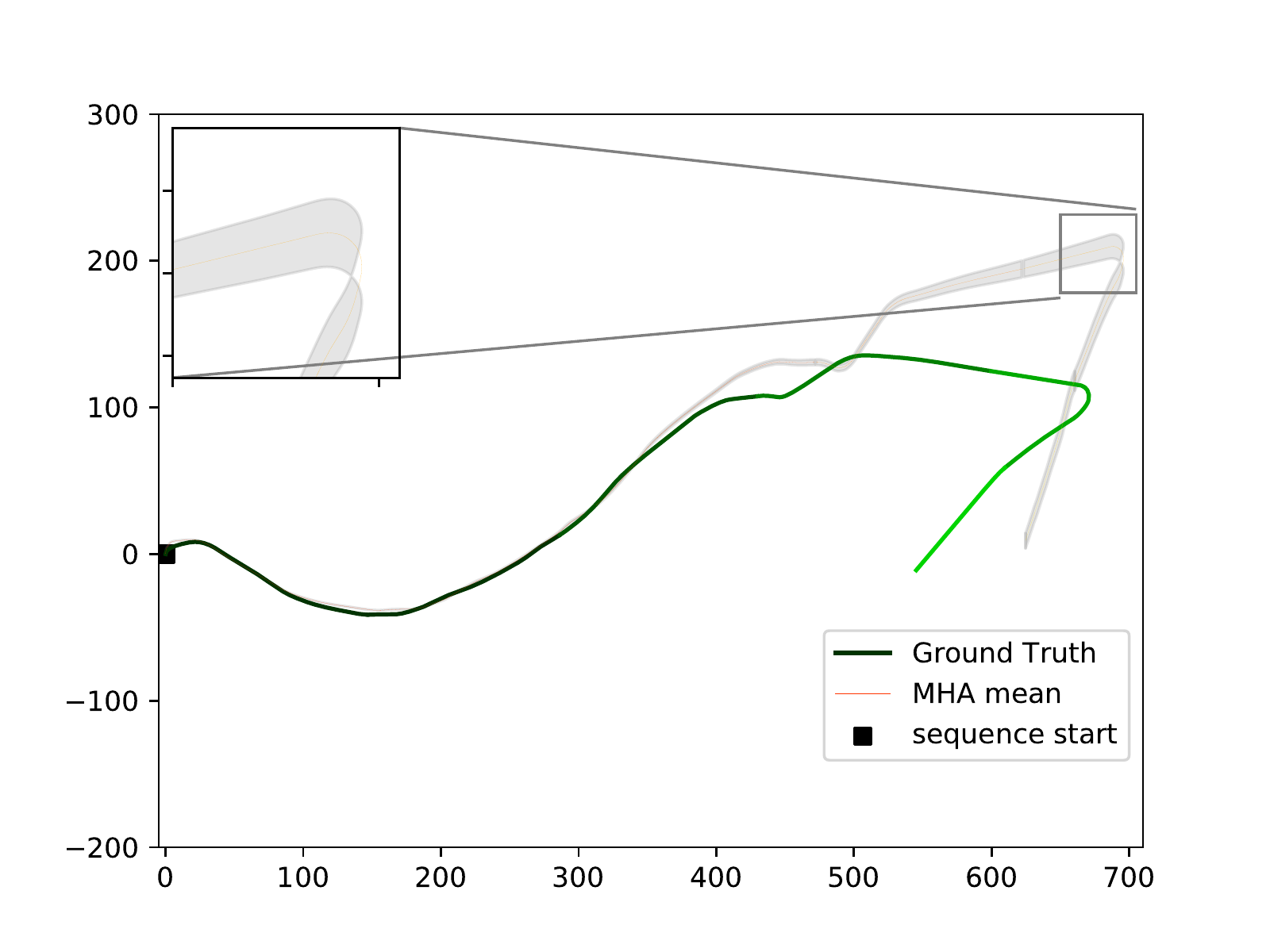}}
\subfigure[Degraded]{\includegraphics[width=40.5mm]{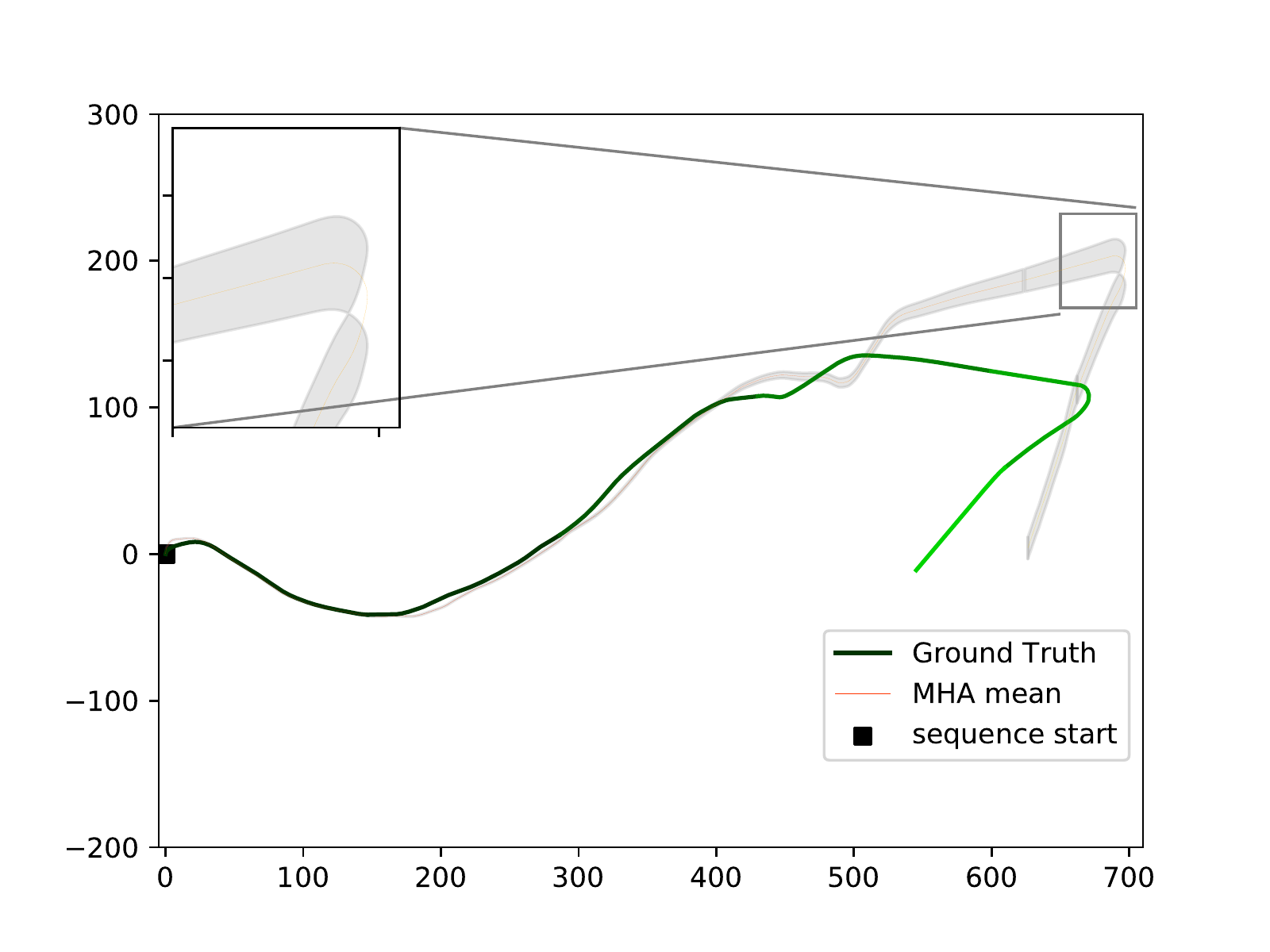}}
\vspace{-2ex}
\caption{Uncertainty in some poses of trajectory for Sequence 10.}\label{fig:10}
\end{figure}

\begin{figure}[ht]
\centering     
\subfigure[]{\includegraphics[width=40.5mm]{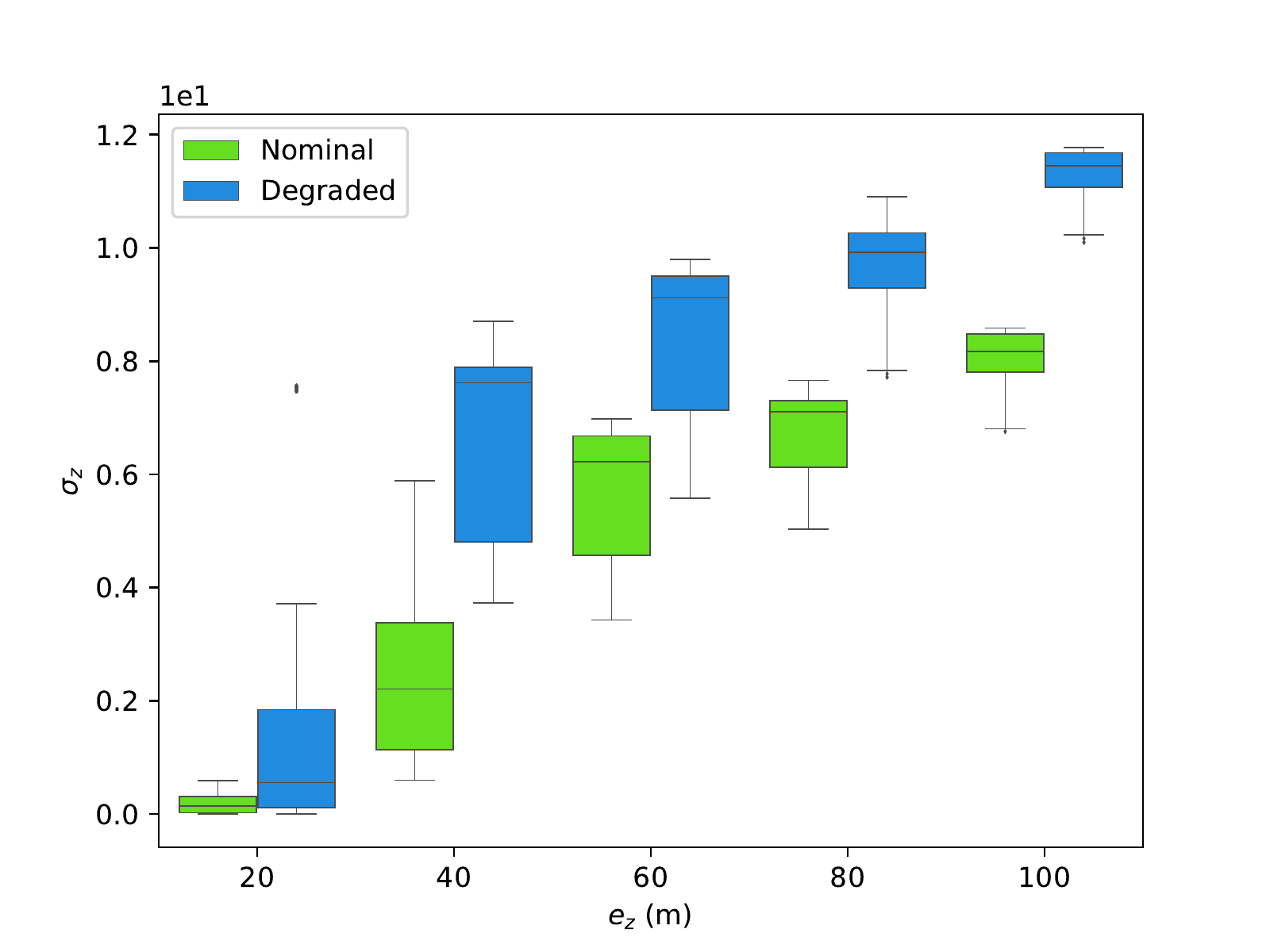}}
\subfigure[]{\includegraphics[width=40.5mm]{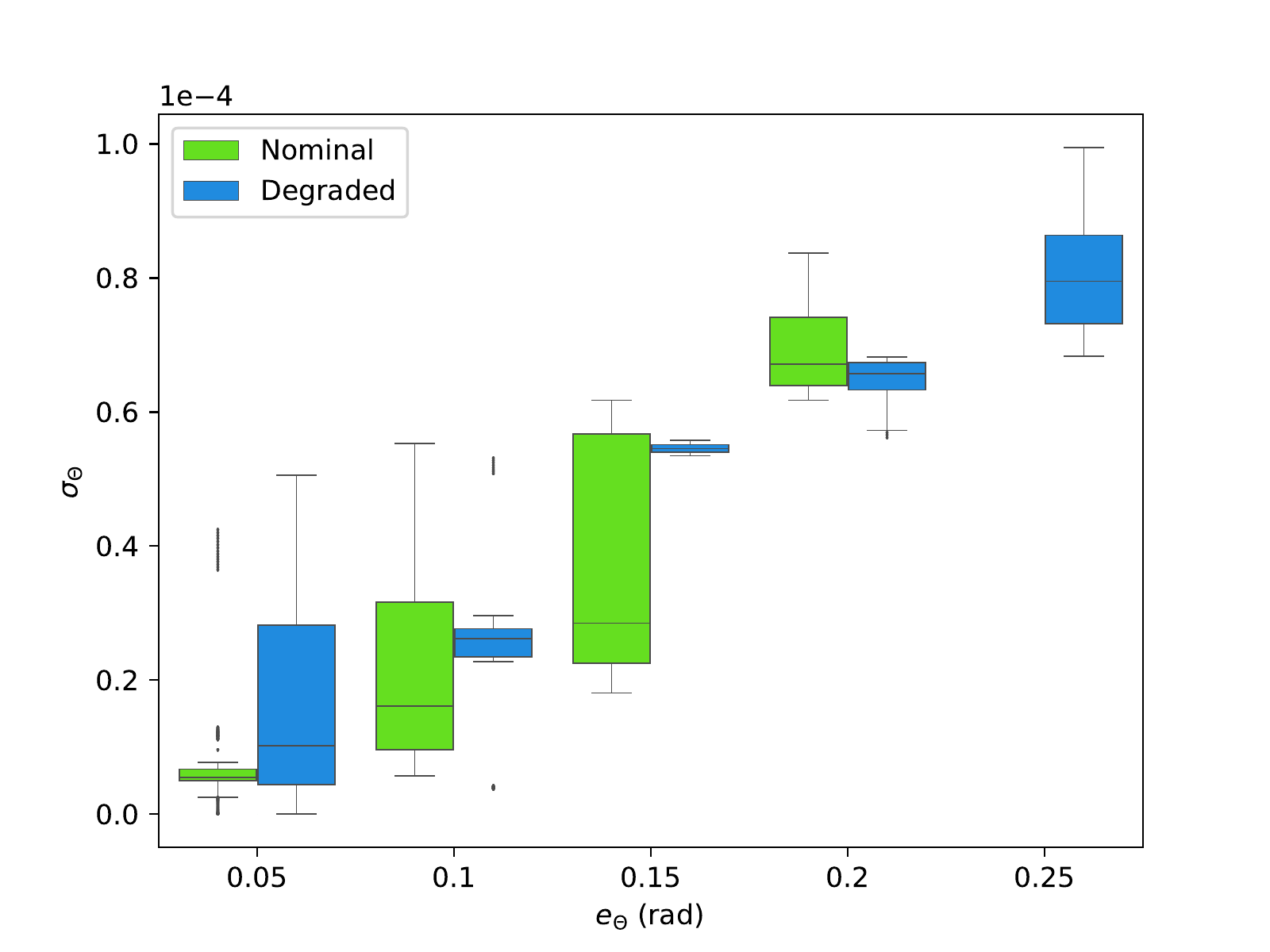}}
\vspace{-2ex}
\caption{Box plots of uncertainty vs. errors for translation along z direction and pitch angle $\theta$ respectively for Sequence 10.}\label{fig:errors}
\end{figure}
Next, we evaluate uncertainty estimates in the form of variance in mean predicted poses, which is depicted with the grey shaded region around the trajectory in Figure~\ref{fig:10}. From Figure~\ref{fig:10}b it can be seen that increased variance expresses the models' uncertainty about the point. To further evaluate the behavioral effects of uncertainty, some more representative results can be seen in Figure~\ref{fig:errors}. The errors and uncertainties are shown for nominal (green) and occlusion degradation (blue) cases. For Sequence \text{10} it was observed that the pose error was large along Z-axis as compared to others. A large uncertainty can be seen along this axis which suggests that the error fluctuations can be better captured by it. The box plots show steady increase in uncertainty along with error (strong correlation with error). Higher uncertainty is observed for degraded case in all plots. Lastly, end-to-end pose regression approaches \citep{posenet} such as ours, are often used in the hybrid methods such as \citep{yang2020d3vo} that effectively leverage the domain knowledge within the learner. In this sense, purely a multi-modal learning approaches can also be useful for such advanced methods beyond end-to-end strategy. For future works, further evaluations on other public datasets \citep{sun2019scalability, nuscenes2019} and other applications of VIO \citep{lutz2020ardea, lee2020visual} can be envisioned.

\section{Conclusion}
We introduced a learning based method for VIO, in which, sensor fusion was carried out using multi-head self-attention. Our approach was evaluated for the scenarios with imperfect sensor data, and we demonstrated that robustness improves with a more expressive fusion strategy. Employing Laplace Approximation, we made our model capable of making predictions with a notion of uncertainty, and demonstrated that the errors in the pose and their confidence can be made correlated for end-to-end VIO based approaches.

\section*{Acknowledgements}
We thank the anonymous reviewers and the area chairs for their time and thoughtful comments. The authors acknowledge the support by the Helmholtz Association’s Initiative and Networking Fund (INF) under the Helmholtz AI platform grant agreement (contract number ID ZT-I-PF-5-1), the project ARCHES (contract number ZT-0033) and the EU-project AUTOPILOT (contract number 731993).


\bibliography{ICML}
\bibliographystyle{icml2020}

\clearpage
\setcounter{section}{0}
\title{\textbf{Supplementary Material}}
\date{}
\maketitle

\section{Related Work}

We consider Deep learning (DL) based approaches for visual localization. Amongst many, we review the most related work as follows. \citet{deepvo} describes an end-to-end trainable approach without using modules from conventional VO pipeline. One of the first deep learning approaches that used CNNs for estimation of 6 DoF poses is PoseNet \cite{posenet}. Some works have shown improved performance by making certain modifications to PoseNet, for e.g. in \citet{Kendall} by changing the loss function to incorporate geometric information. Addition of LSTM after CNN similar to \citet{deepvo} is made to the network in \cite{iccv, vidloc} and the network is further improved by making the CNN part Bayesian to estimate uncertainty of localization in \citet{posenetbayes}. Probabilistic approach has also been explored in \citet{esp}. Several approaches\cite{undeep, unsup3, unsup1, unsup2} have used unsupervised learning approach for depth estimation and VO.

In learning based methods for VIO like VINet \cite{vinet}, wherein the two features streams are simply concatenated and fed into another LSTM for pose regression. It is one of the early end-to-end frameworks for VIO trained in a supervised manner.  The work in \citet{ssf}, replaces the simple fusion strategy used in \citet{vinet} with a more sophisticated one. Two approaches are suggested - soft and hard fusion which do the fusion either deterministic-ally or stocastically. VIOLearner \cite{violearner} is an unsupervised scaled trajectory estimation and online error correction work. However, the necessity of depth information to recover absolute scale may make it difficult to use as it may not always be available. Self-supervised approaches like DeepVIO \cite{deepvio} use optical flow and pre-integrated IMU network with a status update module to update its status, and SelfVIO \cite{selfvio} does monocular VIO and depth reconstruction using adversarial training.

\section{Implementation details}
\subsection{Dataset}
The KITTI odometry benchmark \cite{kitti1, kitti} for odometry evaluations consists of 22 image sequences. Images are acquired at the rate of 10 Hz and saved in png format. OXTS RT 3003 is a GPS/IMU localization system that captures IMU data at 100 Hz (found in raw KITTI dataset) and also determines the ground truth data for this dataset. The ground truth is also acquired at 10 Hz. Sequences 00-10 have corresponding ground truth data associated with them for training whereas sequences 11-22 are without corresponding ground truth for evaluation purposes. We use the sequences 00, 01, 02, 05, 08, 09 for training because they are comparatively longer and sequences 04, 06, 07, 10 for testing. Sequence 03 is omitted due to missing raw data file. Images and raw IMU data are manually synchronized. In order to generate more training data, the sequences are segmented into trajectories of varying lengths which in turn help in avoiding overfitting by core LSTM. Here the varying sequence length is treated as a hyperparameter. 

\subsection{Architecture}
The network is implemented using PyTorch framework \cite {pytorch} and trained on an NVIDIA GTX 1080 GPU. The Adam optimizer with learning rate of 5$\mathrm{e}^{-5}$ is used to train the network for up to 80 epochs with a batch size of 8. The scale factor $\beta$ for MSE is chosen to be 1000 as improved orientation estimates are obtained at this value. Pre-trained FlowNetS model \cite{flownet} is used for the CNN and transfer learning is done from our implementation of the VINet model for both feature extractors to reduce time required for training and convergence. Regularization techniques like batch-normalization and dropout have been used. For the FlowNet adaptation, the CNN has total 17 layers, each convolutional layer followed by ReLU activation except for last convolutional layer. Images have been resized to $184\times 608$. The IMU-LSTM and the core LSTM both are chosen to be bidirectional with two layers with 15 and 1000 hidden states respectively. 

\subsection{Baselines}
For all the baselines, as open source code was not available, and thus, the implementations might differ. The implementation details are as follows. The same configuration for FlowNet and LSTM has been followed for the baselines since a performance improvement was found. specific choice stems For DeepVO, Adagrad optimizer with a learning rate of 5$\mathrm{e}^{-4}$ is used, the network is trained for 250 epochs. VINet is trained using Adam optimizer with learning rate of 1$\mathrm{e}^{-4}$. We have removed the SE(3) composition layer used in original work and replaced it with a fully-connected layer. VIO Soft is trained using Adam optimizer with a learning rate of 1$\mathrm{e}^{-5}$. 

\section{Degradation datasets}
The sensor corruptions tell a lot about network robustness. Robustness of the network to such degradations provides an insight into the underlying sensor fusion mechanism. It points towards the resilience of the network to sensor corruption and consequent failure, highlighting the complementary nature of the sensor modalities in face of adversity. It also gives an insight into how different fusion strategies give relative importance to visual and inertial features. 

For this reason, we prepare two groups of datasets by introducing various types of sensor degradation as done in \cite{ssf}. They are as under:

\subsection{Visual-Inertial degradation}
\begin{itemize}
    \item Part occlusion: We cutout a part of dimension $200\times 200$ pixels from an image at a random location by overlaying a mask of the same size on it. Such situation can occur when camera view is obstructed by objects very close to it or due to dust \cite{occ}.
    \item Noise: We add salt and pepper noise along with Gaussian blur to the images. This can occur due to substantial horizontal motion of the camera, changing light conditions or due to defocusing \cite{blur}. 
    \item Missing frames: Some images are removed at random. This happens when the sensor temporarily gets disconnected or while passing through low-lit area like a tunnel. 
    \item Noise: IMU has inherent errors due to biases in accelerometers and gyroscopes and gyro drift. In addition to the already existing noise we add white noise to accelerometer and bias to gyroscope. This can happen due to temperature variations giving rise to white noise and random walk \cite{imunoise}.
    \item Missing frames: Random removal of IMU frames between two image frames is done. This is plausible due to packet loss from bus or sensor instability. 
\end{itemize}


\section{Additional results for sensor degradation}

\begin{figure*}[h]
\centering    
\subfigure[IMU deg.]{\includegraphics[width=65mm]{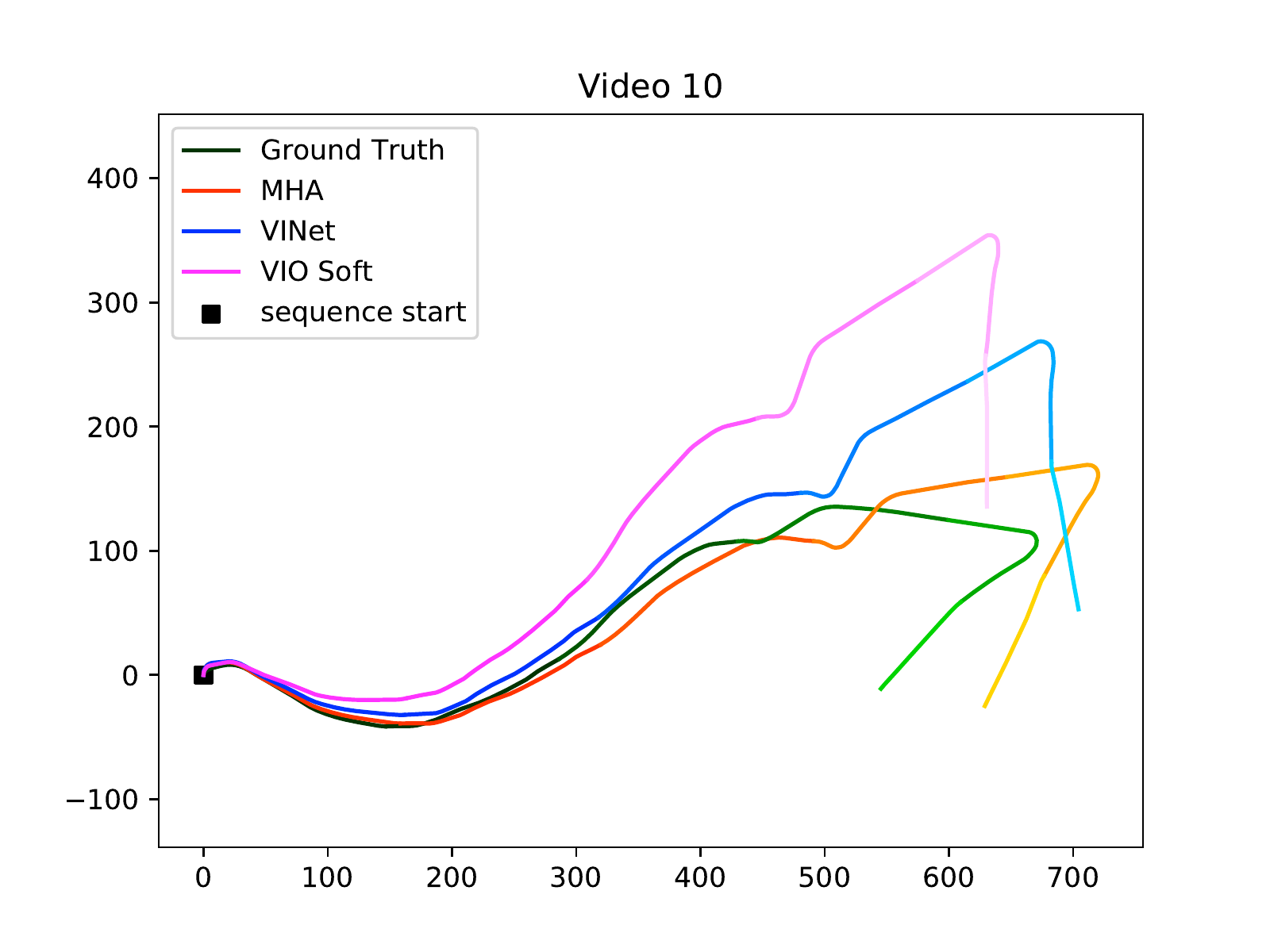}}
\subfigure[All sensor deg.]{\includegraphics[width=65mm]{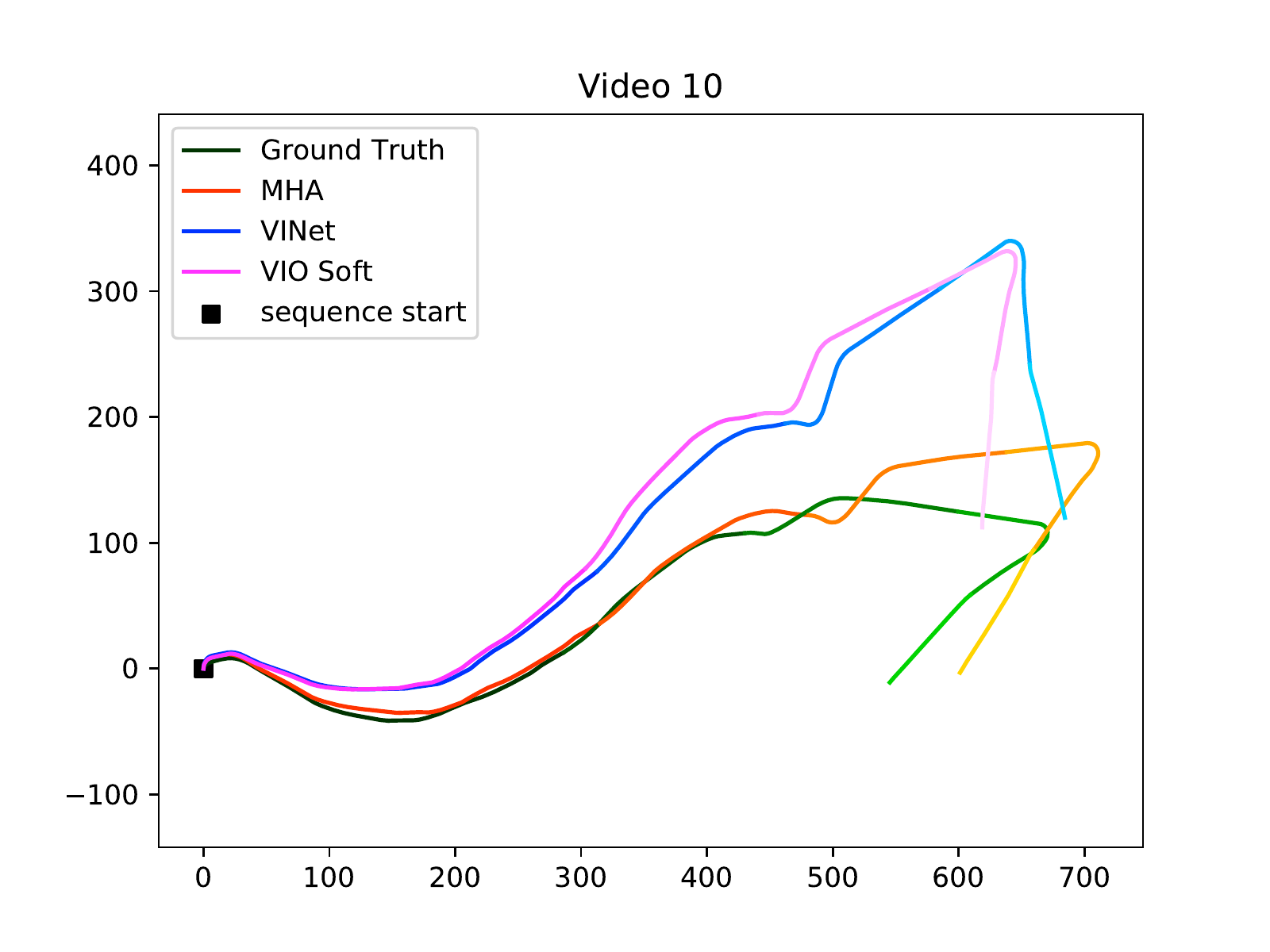}}
\vspace{-2ex}
\caption{Predicted trajectories for (a) IMU degradation (b) all sensor degradation for Sequence 10.}\label{fig:10a}
\end{figure*}

\begin{figure*}[h]
\centering     
\subfigure[Nominal]{\includegraphics[width=65mm]{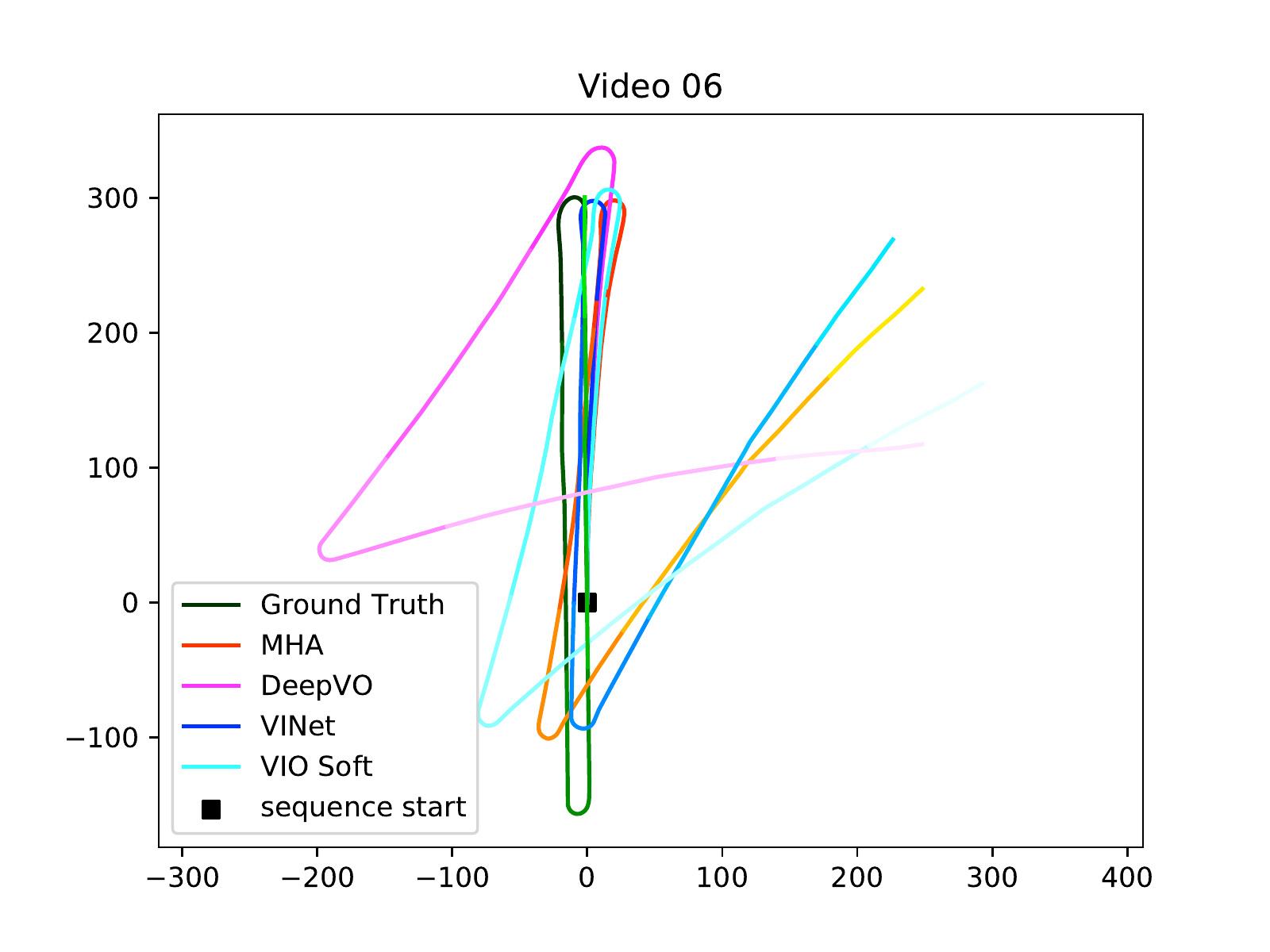}}
\subfigure[IMU deg.]{\includegraphics[width=65mm]{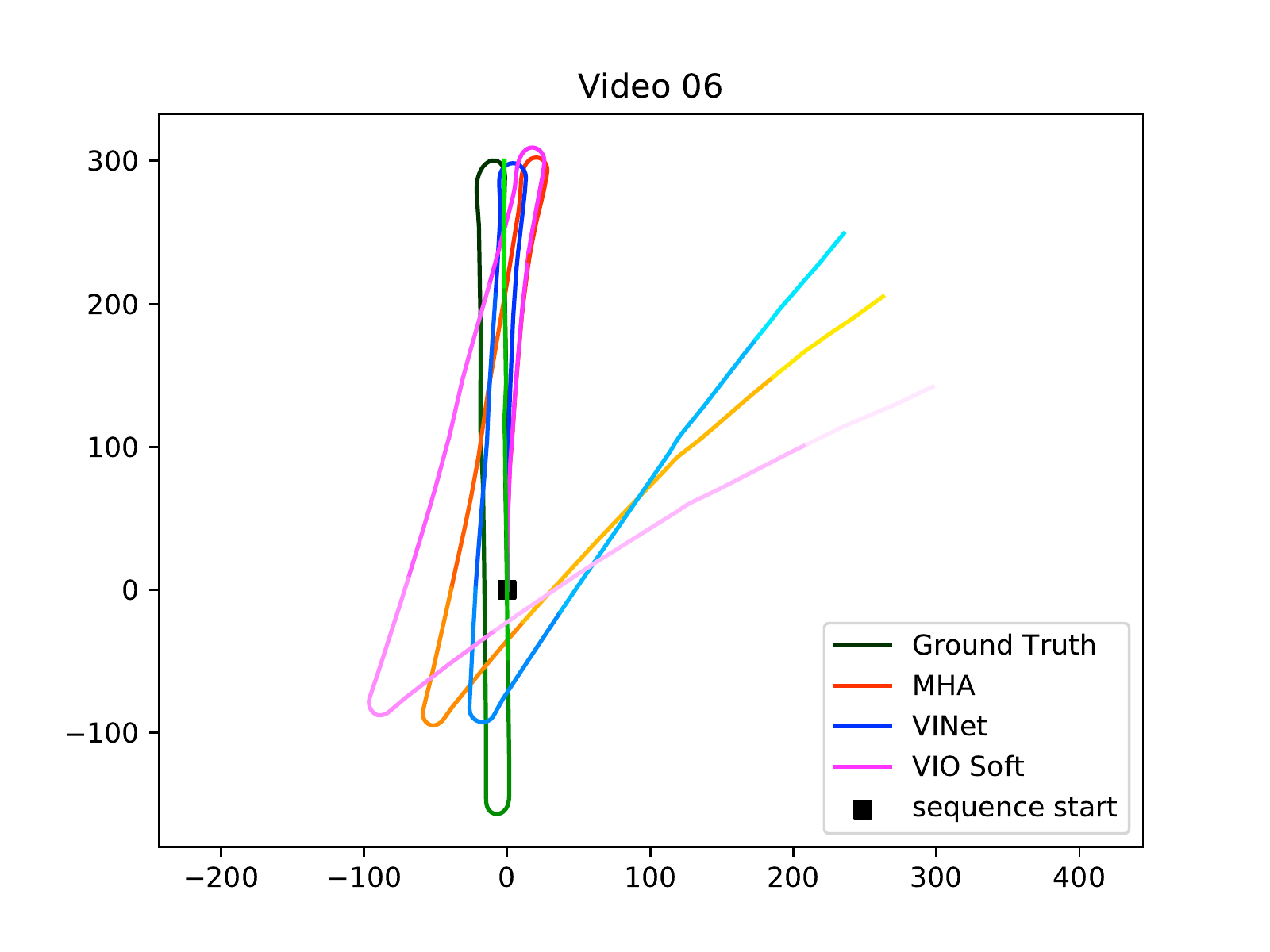}}
\subfigure[Vision deg.]{\includegraphics[width=65mm]{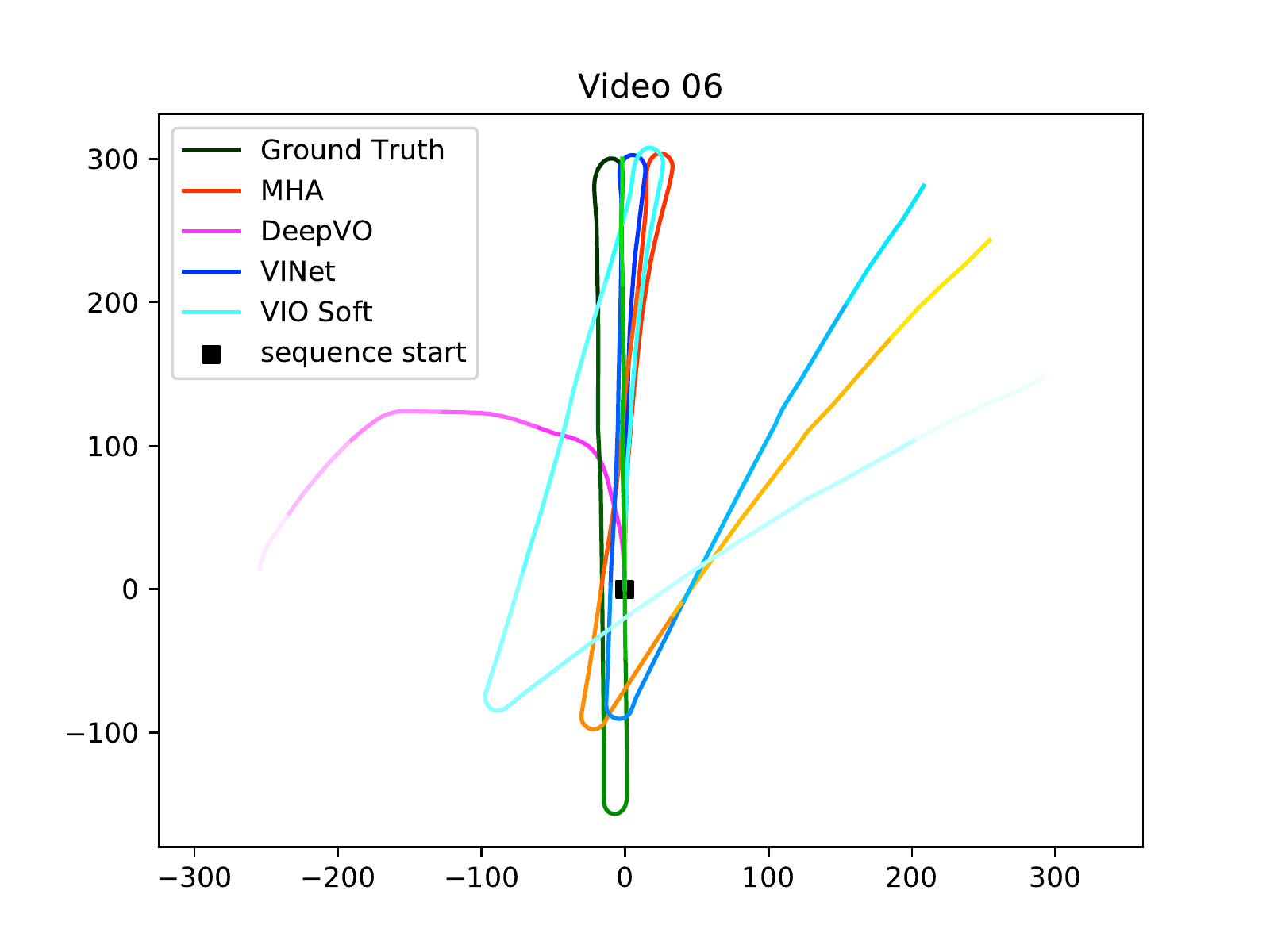}}
\subfigure[All sensor deg.]{\includegraphics[width=65mm]{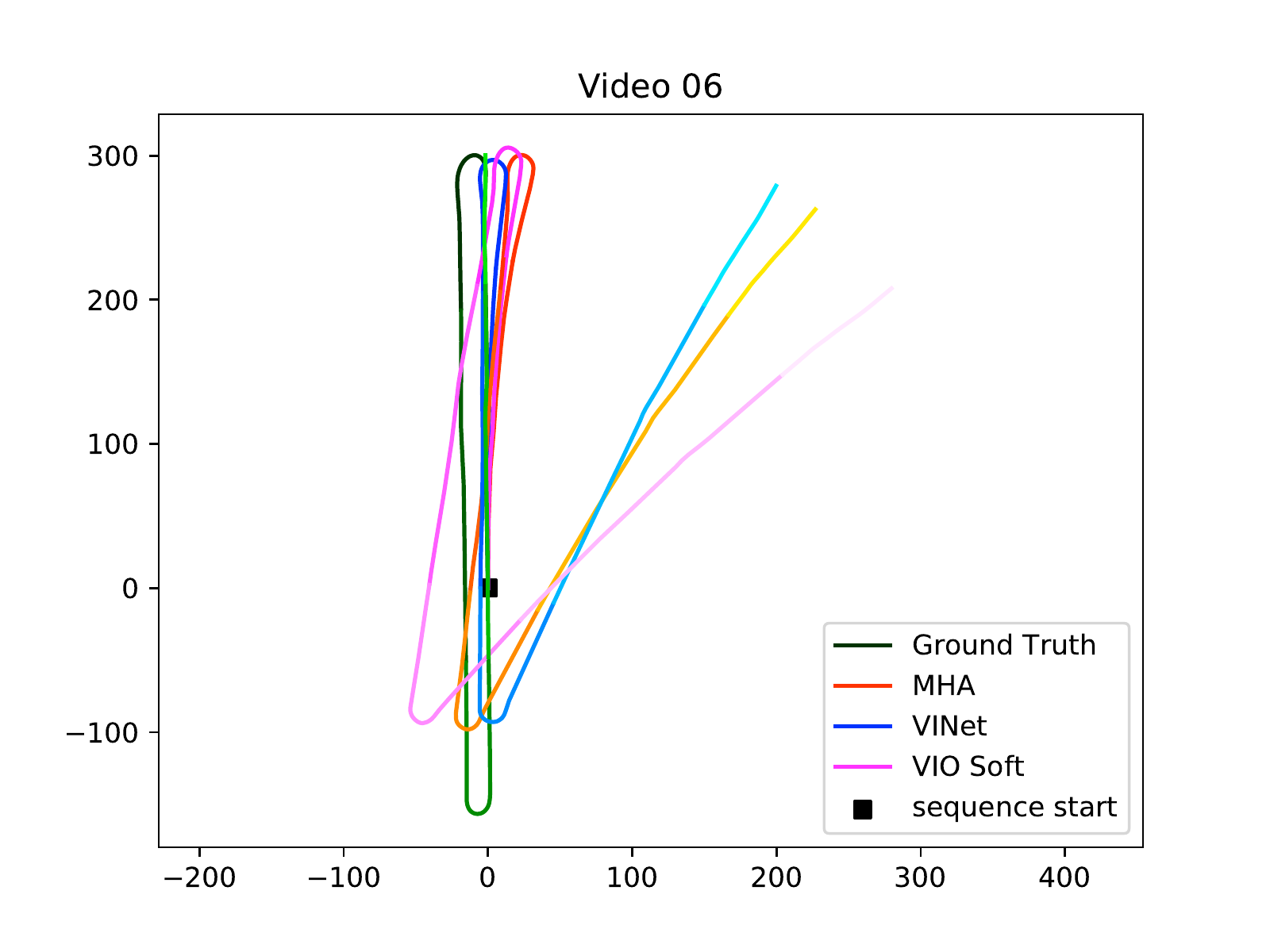}}
\vspace{-2ex}
\caption{Predicted trajectories with corresponding datasets for Sequence 06.}\label{fig:06a}
\end{figure*}

Tables \ref{table:normal}, \ref{table:imu}, \ref{table:allv} and \ref{table:allvi} are the detailed version of Table \ref{table:nom} comparing the metric for sequences 04, 06, 09 and 10. Figure~\ref{fig:10a} and Figure~\ref{fig:06a} substantiate the analysis presented in Section \ref{res1}.

\begin{table}[ht]
\centering
\caption{Comparison metric for nominal case}
\begin{small}
\begin{sc}
\begin{tabulary}{\linewidth}{ CCCCCCCCC }
\toprule
\multirow{2}{*}{Seq} & \multicolumn{2}{c}{DeepVO} & \multicolumn{2}{c}{VINet} & \multicolumn{2}{c}{VIO Soft} & \multicolumn{2}{c}{MHA} \\
\cmidrule(r){2-3}\cmidrule(l){4-5}\cmidrule(l){6-7}\cmidrule(l){8-9}
 & {$t_{rel}$} & {$r_{rel}$} & {$t_{rel}$} & {$r_{rel}$} & {$t_{rel}$} & {$r_{rel}$}  & {$t_{rel}$} & {$r_{rel}$} \\
\midrule
04 & \textbf{6.87} & 2.48 & 11.30 & 2.63 & 10.84 & 2.42 &8.90&\textbf{1.58}\\
06 & 26.52 & 8.90 & \textbf{15.33} & \textbf{5.25} & 17.75 & 6.91&15.59&5.56\\
09 & 18.63 & 4.68 & \textbf{6.95} & \textbf{1.79} & 6.98 & 2.26 &8.38&2.80\\
10 & 23.1 & \textbf{4.33} & 22.34 & 8.98 & 19.03 & 7.74 &\textbf{12.82}&5.80\\
\bottomrule
Mean &18.07&5.88&13.98&4.66&13.65&4.83&\textbf{11.42}&\textbf{3.94} \\
\bottomrule
\label{table:normal}
\end{tabulary}
\end{sc}
\end{small}
\end{table}


\begin{table}[ht]
\centering
\caption{Comparison metric for IMU degradation case}
\begin{small}
\begin{sc}
\begin{tabulary}{\linewidth}{ CCCCCCCCC }
\toprule
\multirow{2}{*}{Seq} & \multicolumn{2}{c}{DeepVO} & \multicolumn{2}{c}{VINet} & \multicolumn{2}{c}{VIO Soft} & \multicolumn{2}{c}{MHA}\\
\cmidrule(r){2-3}\cmidrule(l){4-5}\cmidrule(l){6-7}\cmidrule(l){8-9}
 & {$t_{rel}$} & {$r_{rel}$} & {$t_{rel}$} & {$r_{rel}$} & {$t_{rel}$} & {$r_{rel}$}  & {$t_{rel}$} & {$r_{rel}$} \\
\midrule
04 &N/A&N/A& 12.56 & 3.02 & 10.563 & 2.95 &\textbf{9.30}&\textbf{1.60} \\
06 &N/A&N/A& \textbf{16.71} & \textbf{5.31} & 18.87 & 7.16 &17.04&5.82\\
09 &N/A&N/A& 10.47 & 3.50 & 7.71 & 2.33 &\textbf{7.40}&\textbf{2.13}\\
10 &N/A&N/A& 25.48 & 9.28 & 19.07 & 7.33 &\textbf{16.18}&\textbf{6.15}\\
\bottomrule
Mean &-&-&16.31&5.28&14.05&4.94&\textbf{12.48}&\textbf{3.93} \\
\bottomrule
\label{table:imu}
\end{tabulary}
\end{sc}
\end{small}
\end{table}


\begin{table}[h!]
\centering
\caption{Comparison metric for all vision degradation case}
\begin{small}
\begin{sc}
\begin{tabulary}{\linewidth}{ CCCCCCCCC }
\toprule
\multirow{2}{*}{Seq} & \multicolumn{2}{c}{DeepVO} & \multicolumn{2}{c}{VINet} & \multicolumn{2}{c}{VIO Soft} & \multicolumn{2}{c}{MHA}\\
\cmidrule(r){2-3}\cmidrule(l){4-5}\cmidrule(l){6-7}\cmidrule(l){8-9}
 & {$t_{rel}$} & {$r_{rel}$} & {$t_{rel}$} & {$r_{rel}$} & {$t_{rel}$} & {$r_{rel}$} & {$t_{rel}$} & {$r_{rel}$}  \\
\midrule
04 & F & F & 12.29 & 2.83 & 11.82 & 2.43 &\textbf{9.61}&\textbf{1.90} \\
06 & A & A & \textbf{14.80} & \textbf{5.08} & 19.01 & 6.78 &16.75&5.67\\
09 & I & I & 10.11 & 3.39 & \textbf{7.44} & \textbf{2.20} &10.22&2.84\\
10 & L & L & 24.67 & 8.55 & 20.06 & 6.92 &\textbf{13.86}&\textbf{5.53}\\
\bottomrule
Mean &-&-&15.46&4.97&14.59&4.58&\textbf{12.61}&\textbf{3.98} \\
\bottomrule
\label{table:allv}
\end{tabulary}
\end{sc}
\end{small}
\end{table}

\begin{table}[h!]
\centering
\caption{Comparison metric for all sensor degradation case}
\begin{small}
\begin{sc}
\begin{tabulary}{\linewidth}{ CCCCCCCCC }
\toprule
\multirow{2}{*}{Seq} & \multicolumn{2}{c}{DeepVO} & \multicolumn{2}{c}{VINet} & \multicolumn{2}{c}{VIO Soft} & \multicolumn{2}{c}{MHA}\\
\cmidrule(r){2-3}\cmidrule(l){4-5}\cmidrule(l){6-7}\cmidrule(l){8-9}
 & {$t_{rel}$} & {$r_{rel}$} & {$t_{rel}$} & {$r_{rel}$} & {$t_{rel}$} & {$r_{rel}$} & {$t_{rel}$} & {$r_{rel}$} \\
\midrule
04 &N/A&N/A& 13.59 & 3.09 & 12.98 & 2.61 &\textbf{10.67}&\textbf{2.03} \\
06 &N/A&N/A& \textbf{15.00} & \textbf{4.98} & 16.94 & 5.91 &15.38&5.17\\
09 &N/A&N/A& 10.39 & 3.50 & \textbf{9.34} & 2.71 &9.93&\textbf{2.66}\\
10 &N/A&N/A& 24.49 & 9.00 & 22.65 & 7.28 &\textbf{13.88}&\textbf{5.30} \\
\bottomrule
Mean &-&-&15.87&5.14&15.48&4.63&\textbf{12.47}&\textbf{3.79} \\
\bottomrule
\label{table:allvi}
\end{tabulary}
\end{sc}
\end{small}
\end{table}

\section{Laplace Approximation}
Given model parameters $\theta$ and data $\mathcal{D}$, the log posterior can be captured approximately by a second order Taylor expansion around the maximum a posteriori (MAP) estimate,

\begin{align*}
    \log p(\theta|\mathcal{D}) & \approx \log p(\theta_{MAP}|\mathcal{D}) \\
    & -\frac{1}{2}(\theta-\theta_{MAP})^T H(\theta-\theta_{MAP}),
\end{align*}

where $H$ is the matrix of second order derivatives (Hessian) w.r.t. the negative log posterior. Notably, the right hand side is of Gaussian functional form when applying the exponential, thus it can be used to sample parameter configurations:

\[
  \theta\sim\mathcal{N}(\theta_{MAP},H^{-1}).
\]

Because the Hessian of a deep neural network is infeasible to compute and invert, approximations to the Fisher information matrix (Fisher), like Kronecker factored or diagonal approximations, are used instead.

Approximate Bayesian inference can be performed using Monte Carlo integration, by sampling multiple parameter configurations from the approximate posterior and collecting the model output for each of them given an unseen datum $\mathbf{\tilde x}$:

\begin{align*}
    p(\mathbf{\tilde y}|\mathbf{\tilde x},\mathcal{D})&=\int p(\mathbf{\tilde y}|\mathbf{\tilde x},\theta)p(\theta|\mathcal{D})\mathrm{d}\theta\\
    &\approx\frac{1}{T}\sum_{t=1}^T p(\mathbf{\tilde y}|\mathbf{\tilde x},\theta_t).
\end{align*}

The mean is used to make predictions while for regression, the variance is a measure of the predictive uncertainty. We make use of $T=30$ posterior samples in all our experiments. Because of numerical instability, overestimation of variance and wrong attribution of probability mass, the posterior approximation needs to be regularized in practice. We make use of the following simple regularization scheme:

\[
  NF+\tau I,
\]

where $F$ is the approximation to the Fisher and $N$ and $\tau$ are treated as hyperparameters to be optimized on a validation set. We made use of our own implementation of LA for Bayesian inference in PyTorch which we will make publicly available in \url{https://github.com/DLR-RM}.

\end{document}